\documentclass[]{informs4anon}

\usepackage{fancyvrb}
\usepackage{hyperref}
\usepackage{mathtools}

\usepackage{natbib}
 \bibpunct[, ]{(}{)}{,}{a}{}{,}%
 \def\bibfont{\small}%

\OneAndAHalfSpacedXII
\TheoremsNumberedThrough
\EquationsNumberedThrough

\MANUSCRIPTNO{}

\usepackage{xspace}
\newcommand{\MathOptAI}[0]{\texttt{MathOptAI.jl}\xspace}
\newcommand{\gurobiml}[0]{\texttt{gurobi-machinelearning}\xspace}
\newcommand{\OMLT}[0]{\texttt{OMLT}\xspace}
\newcommand{\pyscipoptml}[0]{\texttt{PySCIPOpt-ML}\xspace}
\newcommand{\gamspy}[0]{\texttt{GAMSPy}\xspace}
\newcommand{\isvec}[1]{\textbf{#1}}

\newcommand{\revision}[1]{#1}
\newcommand{\revisionb}[1]{#1}

\begin{document}


\RUNAUTHOR{Dowson, Parker, and Bent}
\RUNTITLE{MathOptAI.jl: Embed trained machine learning predictors into JuMP models}

\TITLE{MathOptAI.jl: Embed trained machine learning predictors into JuMP models}

\ARTICLEAUTHORS{%
\AUTHOR{Oscar Dowson}
\AFF{Dowson Farms, Auckland 1061, New Zealand, \EMAIL{oscar@dowsonfarms.co.nz}}
\AUTHOR{Robert B.~Parker}
\AFF{Los Alamos National Laboratory, Los Alamos, NM 87544, USA, \EMAIL{rbparker@lanl.gov}}
\AUTHOR{Russell Bent}
\AFF{Los Alamos National Laboratory, Los Alamos, NM 87544, USA, \EMAIL{rbent@lanl.gov}}
}

\ABSTRACT{
We present \MathOptAI, an open-source Julia library for embedding trained machine learning predictors into a JuMP model. \MathOptAI can embed a wide variety of neural networks, decision trees, and Gaussian Processes into a larger mathematical optimization model. In addition to interfacing a range of Julia-based machine learning libraries such as \texttt{Lux.jl} and \texttt{Flux.jl}, \MathOptAI uses Julia's Python interface to provide support for PyTorch models. When the PyTorch support is combined with \MathOptAI's gray-box formulation, the function, Jacobian, and Hessian evaluations associated with the PyTorch model are offloaded to the GPU in Python, while the rest of the nonlinear oracles are evaluated on the CPU in Julia.
}


\maketitle

\section{Introduction}

A recent trend in the mathematical optimization literature is to embed trained machine learning predictors into a larger optimization model. The most common application is for a practitioner to train a machine learning predictor as a surrogate for a more complicated subsystem that cannot be directly embedded into an optimization model, for example, because it does not have an algebraic form or because it is non-differentiable.
\citet{lopezflores2024} provide a review of the field.

We present \MathOptAI, an open-source Julia \citep{bezanson_julia_2017} library for embedding trained machine learning predictors into a JuMP \citep{jump1} model. \MathOptAI can embed a wide variety of neural networks, decision trees, and Gaussian processes into a larger mathematical optimization model. \MathOptAI supports a range of mixed-integer linear, smooth nonlinear, and non-smooth nonlinear reformulations. The design of \MathOptAI is inspired by previous work such as \gurobiml \citep{gurobimachinelearning} and \OMLT \citep{omlt}.

More formally, the purpose of \MathOptAI is to build optimization problems in the form given by
Problem \eqref{eqn:nlopt}:
\begin{equation}
  \begin{array}{cl}
   \displaystyle \min_{\isvec{x}\in\mathbb{R}^N,\ \isvec{y}\in\mathbb{R}^P} & f_0(\isvec{x},\isvec{y}) \\
   \text{s.t.} & f_i(\isvec{x},\isvec{y}) \in S_i,\quad i = 1,\ldots,M\\
               & \isvec{y} = F(\isvec{x}), \\ 
  \end{array}
  \label{eqn:nlopt}
\end{equation}
where $f_i(\isvec{x}, \isvec{y}): \mathbb{R}^{N+P} \rightarrow \mathbb{R}^{D_i}$ is a function that must belong to the set $S_i\subset \mathbb{R}^{D_i}$. \revision{Note that, here and throughout, bold-faced symbols are vectors, and $x_i$ is the $i$-th component of the vector $\isvec{x}$.} The functions $f_i$ and sets $S_i$ are defined by the MathOptInterface standard form \citep{legat2021mathoptinterface} and abstract over the various problem classes of mathematical optimization, from mixed-integer linear to nonlinear. The relationship $\isvec{y} = F(\isvec{x})$, where $F(\isvec{x}): \mathbb{R}^N \rightarrow \mathbb{R}^P$, is the machine learning predictor (or concatenation of multiple predictors) that we want to embed in the optimization model. We delay explaining how we embed the relationship $\isvec{y} = F(\isvec{x})$ until Section~\ref{sec:algebraic}.

\subsection{Examples of machine learning predictors}

Examples of predictors $F(\isvec{x})$ supported by \MathOptAI include:
\begin{itemize}
    \item $\textsf{Affine}_{\{A,\isvec{b}\}}(\isvec{x}) = A\isvec{x} + \isvec{b}$
    \item $\textsf{ReLU}(\revision{\isvec{x}}) = \max(\revision{\isvec{0}}, \revision{\isvec{x}})$
    \item $\textsf{Sigmoid}(\revision{\isvec{x}}) = \frac{1}{1 + e^{-\revision{\isvec{x}}}}$
    \item $\textsf{SoftMax}(\isvec{x}) = \frac{e^\isvec{x}}{||e^\isvec{x}||_1}$
    \item $\textsf{Tanh}(\revision{\isvec{x}}) = \tanh(\revision{\isvec{x}})$.
\end{itemize}
\revision{By convention, all predictors take vector inputs and produce vector outputs. Some predictors are naturally vector-valued, for example, $\textsf{Affine}_{\{A,\isvec{b}\}}(\isvec{x}): \mathbb{R}^m \rightarrow \mathbb{R}^n$, where $m$ and $n$ are derived from the shape of $A$ and $\isvec{b}$. Other predictors involve operators (like $e$, $\max$, and $\tanh$) that produce scalar outputs. When scalar-valued operators are applied to a vector input, the convention is that the operator is applied element-wise.}

\MathOptAI also supports predictors which are not simple algebraic equations. One example is a binary decision tree ($\mathbb{R}^m \rightarrow \mathbb{R}^{\revision{1}}$), which can be represented by the set of constraints:
\begin{align*}
    \textsf{BinaryDecisionTree}_{\{\mathcal{P},f,y\}}(\isvec{x}) = & \sum\limits_{p\in \mathcal{P}} \delta_p y_p \\
    \text{such that:\;\;\;}& \delta_p \implies f_n(\isvec{x}) \le 0 & \forall p\in\mathcal{P}, n \in p \\
    & \sum\limits_{p\in \mathcal{P}} \delta_p=1 \\
    & \delta_p \in \{0, 1\} & \forall p \in \mathcal{P},
\end{align*}
where $\mathcal{P}$ is a set of paths through the tree, $y_p$ is the leaf value associated with path $p$, and $f_n$ describes the relation that must hold at each node $n$ if path $p$ is taken \citep{gurobimachinelearning}.

Some predictors are defined by the composition of simpler predictors. For example, a logistic regression predictor is:
\begin{equation*}
    \isvec{y} = F(\isvec{x}) = \frac{1}{1 + e^{-(A\isvec{x} + \isvec{b})}},
\end{equation*}
which is the composition:
$$\isvec{y} = F(\isvec{x}) = (\textsf{Sigmoid} \circ \textsf{Affine}_{\{A,\isvec{b}\}})(\isvec{x}).$$

Similarly, a feed-forward neural network predictor is defined by the repeated application of an affine transformation and a nonlinear activation function $\sigma$ over $L$ layers:
\begin{equation*}
  \begin{aligned}
    \isvec{y}_l &= \sigma_l (W_l \isvec{y}_{l-1} + \isvec{b}_l), \quad l =1,\dots,L,
  \end{aligned}
\end{equation*}
where $\isvec{y}_0 = \isvec{x}$ and $\isvec{y} = \isvec{y}_L$. Examples of activation functions include $\sigma(x) = \max(0, x)$, and $\sigma(x) = \tanh(x)$. Instead of considering a layer comprised of both an affine transform and an activation function, \MathOptAI models a neural network as the sequential composition of predictors. For example:
$$\isvec{y} = F(\isvec{x}) = (\textsf{SoftMax} \circ \textsf{Affine}_{\{W_L,\isvec{b}_L\}} \circ \cdots \circ \textsf{ReLU} \circ \textsf{Affine}_{\{W_1,\isvec{b}_1\}})(\isvec{x}).$$

In addition to the predictors listed above, \MathOptAI also supports more complicated predictors, such as random forests, gradient boosted trees, and a predictor that returns the quantile of a distribution or Gaussian Process. We omit these formulations in the interest of brevity. Moreover, due to the on-going open-source development of the package, we also omit a complete list of the supported predictors since the list will inevitably become outdated. Readers should consult the \revision{associated code archive \citep{DowsonMathOptAI}} for the full list of supported predictors.

\subsection{Contribution and Outline}

The main contribution of this paper is the Julia library \MathOptAI, \revision{an open-source Julia \citep{bezanson_julia_2017} library for embedding trained machine learning predictors into a JuMP \citep{jump1} model. The \MathOptAI library is archived at \cite{DowsonMathOptAI}.}

A novel feature of \MathOptAI is its support for the gray-box formulation of a neural network (discussed in Section~\ref{sec:gray-box}). \revision{Most usefully, the gray-box formulation allows us to embed large-scale neural networks into a Julia-based JuMP model whilst evaluating the associated zeroth-, first-, and second-order derivatives of the network's loss function on a GPU from Python.} Although we are not the first to develop the technique (see \cite{casas2024} and \cite{casas2025comparison}), we are the first to present it in a general purpose library. \revision{As secondary contributions, we provide a review of software alternatives to \MathOptAI, and we explain our design principles. These contributions may be useful for authors looking to implement a similar package in other programming languages.}

Readers should note that the purpose of this paper is not to provide computational experiments of the relative merits of embedding machine learning predictors into an optimization model. We direct interested readers to our related work \citep{parker2024formulationsscalabilityneuralnetwork}, and to the ever growing literature on the subject like \cite{casas2025comparison}.

The rest of this paper is organized as follows. In Section~\ref{sec:algebraic}, we explain how we embed a predictor into an optimization model. In Section~\ref{sec:example} we provide a code example of using \MathOptAI. Section~\ref{sec:comparison} is a comparison to related work, and we conclude in Section~\ref{sec:principles} with our guiding design principles.

\section{Algebraic representation}\label{sec:algebraic}

This section explains the three ways in which \MathOptAI embeds the $\isvec{y} = F(\isvec{x})$ equation into an optimization model. The three approaches are denominated \textit{full-space}, \textit{reduced-space}, and \textit{gray-box}.

\subsection{Full-space}

In the \textit{full-space} formulation, we add a vector-valued decision variable $\isvec{y}$ to represent the output of each predictor, and we add constraints and additional variables as necessary to ensure that $\isvec{y} = F(\isvec{x})$ holds in a feasible solution. For example, an $\textsf{Affine}_{\{A,\isvec{b}\}}$ predictor is encoded as the vector-valued linear equality constraint:
\begin{equation*}
    \revision{\isvec{y} = A\isvec{x} + \isvec{b}},
\end{equation*}
and a $\textsf{Sigmoid}$ predictor is encoded as the set of nonlinear equality constraints: 
\begin{equation*}
    y_i = \frac{1}{1 + e^{-x_i}},\quad i = 1,\ldots,P,
\end{equation*}
where $P$ is the output dimension of the predictor.

Predictors formed by the composition of other predictors re-use the intermediate $y_i$ variables at each step of the composition. For example, the logistic regression example $\textsf{Sigmoid} \circ \textsf{Affine}_{\{A,\isvec{b}\}}$ can be formulated as:
\begin{equation*}
\begin{aligned}
    \revision{\isvec{z} = A\isvec{x} + \isvec{b}} \\
    \revision{y_{i} = \frac{1}{1 + e^{-z_i}}}&,\quad i = 1,\ldots,P_1,
\end{aligned}
\end{equation*}
where $P_1$ is the output dimension of the affine predictor.

Some formulations may introduce additional variables and constraints. For example, the \textit{complementarity} or \textit{quadratic} formulation of $\textsf{ReLU}$ is:
\begin{equation*}
\begin{aligned}
    y_i = x_i + z_i,\quad i = 1,\ldots,P \\
    y_i \cdot z_i \le 0,\quad i = 1,\ldots,P \\
    y_i, z_i \ge 0,\quad i = 1,\ldots,P.
\end{aligned}
\end{equation*}

The benefits of the \textit{full-space} approach are that solvers and modeling languages can efficiently exploit the affine constraints, and that each nonlinear constraint is sparse (it uses only a small number of the total decision variables) and involves few nonlinear terms. The downside of the \textit{full-space} approach is that we add intermediate variables and constraints. For deep (or wide) neural networks in particular, this can result in a large number of additional variables and constraints in the optimization model. Because of its simplicity, the \textit{full-space} formulation is supported by all software packages that we compare in Section~\ref{sec:comparison}.

\subsection{Reduced-space}
In the \textit{reduced-space} formulation we represent each predictor as an expression, and we delay adding intermediate variables until necessary. For example, the logistic regression example is equivalent to the expression:
\begin{equation*}
    y_{i} \coloneq \frac{1}{1 + e^{\revision{-\isvec{e}_i^\top(A\isvec{x} + \isvec{b})}}},\quad i = 1,\ldots,P_1.
\end{equation*}
If $y_i$ appears in other parts of the optimization model, then the nonlinear expression is used instead of an intermediate decision variable $y_i$.

Note that some predictors do not have a reduced-space formulation. One example is the complementarity $\textsf{ReLU}$ formulation. For these predictors we always use the full-space formulation.

The benefit of the \textit{reduced-space} approach is that we add fewer variables and constraints to the problem. The downside is that the nonlinear expressions become complicated with a very large number of terms. Of the packages that we compare in Section~\ref{sec:comparison}, the \textit{reduced-space} formulation is supported only by \MathOptAI and \OMLT. One reason is that other software packages like \gurobiml have limited support for large nonlinear expressions.

\subsection{Gray-box}\label{sec:gray-box}

In the \textit{gray-box} formulation, we do not attempt to encode the predictor $F$
algebraically. Instead, we exploit the fact that nonlinear local solvers such as Ipopt \citep{ipopt}
require only callback oracles to evaluate the function $F(\isvec{x}): {\mathbb{R}}^N \rightarrow \mathbb{R}^P$, the Jacobian
$\nabla F(\isvec{x}): {\mathbb{R}}^N \rightarrow \mathbb{R}^{P\times N}$ and,
optionally, the Hessian-of-the-Lagrangian $\nabla^2\mathcal{L}(\isvec{x}, \lambda): {\mathbb{R}}^{N+P}
\rightarrow \mathbb{R}^{N\times N}$, which is defined as:
$$\nabla^2\mathcal{L}(\isvec{x}, \lambda) = \sum\limits_{i=1}^P \lambda_i \nabla^2 F_i(\isvec{x}),$$
where $\lambda \in \mathbb{R}^P$ is a vector of Lagrange multipliers that varies between iterations of the solution algorithm.

Using JuMP's support for user-defined nonlinear operators, we implement the function evaluation as a nonlinear operator $F(\isvec{x})$, and we use automatic differentiation to compute the Jacobian, $\nabla F(\isvec{x})$, and Hessian-of-the-Lagrangian, $\nabla^2\mathcal{L}(\isvec{x}, \lambda)$.  Then, we add $F$ using the full-space formulation:
\begin{equation*}
    F(\isvec{x}) - \isvec{y} = \isvec{0},
\end{equation*}
or the reduced-space formulation $\isvec{y} \coloneq F(\isvec{x})$.

There are three main benefits to the \textit{gray-box} approach. First, the number of variables and constraints in the optimization model scales with the dimension of the input vector $\isvec{x}$ and output vector $\isvec{y}$, and not with the size or complexity of the intermediate predictors because these are not represented explicitly in the optimization model. As shown in our related work \citep{parker2024formulationsscalabilityneuralnetwork}, this enables the \textit{gray-box} formulation to scale and embed machine learning predictors with hundreds of millions of parameters. 
Second, the external evaluation of the oracles means we can trivially use customized tools, such as PyTorch's GPU acceleration, to improve the evaluation performance. We again direct readers to \cite{parker2024formulationsscalabilityneuralnetwork} for more details.
Third, the gray-box formulation enables models which are too complicated (or tedious) to express algebraically in the full-space or reduced-space formulations. This includes, for example, user-defined custom layers in PyTorch.

The downside of the \textit{gray-box} approach is that the explicit representation of the constraints are not exposed to the solver, which means that the gray-box formulation cannot be used by nonlinear {\it global} optimization solvers. Moreover, some predictors do not have a gray-box formulation because they are not amenable to oracle-based nonlinear optimization. One example is the \textsf{BinaryDecisionTree} predictor, which is discrete and non-differentiable.

Of the packages that we compare in Section~\ref{sec:comparison}, the \textit{gray-box} formulation is unique to \MathOptAI.
The \textit{gray-box} formulation was previously suggested by \cite{casas2024} and \cite{casas2025comparison}, but they developed their work as a custom implementation and not as part of a wider general purpose library.

\subsubsection{Hessian-of-the-Lagrangian}

Naively computing the Hessian-of-the-Lagrangian for the \textit{gray-box} formulation requires computing $P$ many $N\times N$ Hessian matrices $\nabla^2 F_i(\isvec{x},\lambda)$, and then computing a sum-product over the $\lambda$ vector (in $\mathbb{R}^P$).
We improve performance by creating the Lagrangian function as a new predictor:
$$G(\isvec{x}, \lambda) = (\textsf{Affine}_{\{\lambda,\isvec{0}\}} \circ F)(\isvec{x}),$$
where $\lambda$ is treated as a $1\times P$ matrix, then differentiating $G$ directly using the automatic differentiation frameworks of PyTorch \citep{paszke2019pytorch} or \texttt{Flux.jl} \citep{Innes2018}.
Integrating with machine learning frameworks to evaluate the Hessian-of-the-Lagrangian allows us to avoid the sum-product and memory requirement to represent the explicit Hessians, and to exploit GPU acceleration, if available, when computing $\nabla^2 G(\isvec{x},\lambda)$.

\subsection{Relative performance}

\revision{The relative performance of each formulation depends on the problem
instance, solution algorithm, and machine hardware.
For example, the full-space formulation may perform better on some instances
due to the advantages of ``lifting'' discussed by \cite{Albersmeyer}.
In contrast, we have observed that reduced-space formulations sometimes
require significantly fewer iterations to converge
\citep{parker2024formulationsscalabilityneuralnetwork}.
While gray-box formulations can exploit GPU acceleration, we note that similar
speedups may be possible for full-space formulations by exploiting the structure
of the machine learning model in the solver's linear algebra operations.
We leave a numerical comparison of the different approaches to future work.}

\subsection{Variable bounds}

In addition to adding the variables and constraints required by each predictor, we also add bounds to the new variables by propagating bounds from the input $\revision{\isvec{x}}$ variables (when they exist) through the predictors by direct substitution. Although theoretically redundant, adding these bounds often improves the performance of the optimization solver.

\section{Code example}\label{sec:example}

As a trivial example of the usage of \MathOptAI, we consider embedding the following neural network into a JuMP model. It has 10 input variables, two output variables, and a $\textsf{ReLU}$ activation function. 

First, we build and train (code not shown) the model in PyTorch \citep{paszke2019pytorch} . Then we save the trained model to disk using \texttt{torch.save}:

\begin{Verbatim}[fontsize=\small, frame=lines]
#!/usr/bin/python3
import torch
from torch import nn
model = nn.Sequential(nn.Linear(10, 16), nn.ReLU(), nn.Linear(16, 2))
torch.save(model, "model.pt")
\end{Verbatim}
We can then embed the network from PyTorch into a JuMP model:
\begin{Verbatim}[fontsize=\small, frame=lines]
#!/usr/bin/julia
using JuMP, MathOptAI, PythonCall, Ipopt
model = Model(Ipopt.Optimizer)
@variable(model, 0 <= x[1:10] <= 1)
predictor = MathOptAI.PytorchModel("model.pt")
y, formulation = MathOptAI.add_predictor(model, predictor, x)
\end{Verbatim}
These lines: load the relevant Julia packages; create a JuMP model with Ipopt \citep{ipopt} as the optimizer; define a vector of 10 decision variables; reference the PyTorch file on disk; and add the predictor using the \textit{full-space} formulation. The return \texttt{y} is the output vector of decision variables.
The \texttt{formulation} object contains the variables and constraints added by the predictor. To add the predictor using the reduced-space formulation, we can do:
\begin{Verbatim}[fontsize=\small, frame=lines]
y, _ = MathOptAI.add_predictor(model, predictor, x; reduced_space = true)
\end{Verbatim}
To add the predictor using the gray-box formulation, we can do:
\begin{Verbatim}[fontsize=\small, frame=lines]
y, _ = MathOptAI.add_predictor(model, predictor, x; gray_box = true)
\end{Verbatim}
The gray-box formulation has additional options. For example, here we evaluate the oracles on the GPU (\texttt{device = "cuda"}) and we do not provide Hessians to Ipopt (\texttt{hessian = false}):
\begin{Verbatim}[fontsize=\small, frame=lines]
y, _ = MathOptAI.add_predictor(model, predictor, x; gray_box = true,
                               device = "cuda", hessian = false)
\end{Verbatim}

By default, \MathOptAI uses the non-smooth nonlinear formulation of $\textsf{ReLU}$. To support a mixed-integer linear solver like HiGHS \citep{huangfu_2018}, we can specify a different reformulation:

\begin{Verbatim}[fontsize=\small, frame=lines]
#!/usr/bin/julia
using JuMP, MathOptAI, PythonCall, HiGHS
model = Model(HiGHS.Optimizer)
@variable(model, 0 <= x[1:10] <= 1)
predictor = MathOptAI.PytorchModel("model.pt")
config = Dict(:ReLU => MathOptAI.ReLUSOS1)
y, _ = MathOptAI.add_predictor(model, predictor, x; config)
\end{Verbatim}
Here \texttt{config} is a dictionary which maps the name of a PyTorch layer to an \texttt{AbstractPredictor} \revision{constructor} for \MathOptAI to use when formulating the predictor. The \texttt{ReLUSOS1} formulation uses the SOS1 formulation of \cite{turner2023pyscipopt}.
Note that HiGHS does not support SOS1 constraints natively, but, when the input variables are bounded, JuMP automatically reformulates SOS1 constraints into a MIP equivalent.

\MathOptAI supports a variety of machine learning predictors that are not neural networks. For example, to train and embed a decision tree from \texttt{DecisionTree.jl} \citep{ben_sadeghi_2022_7359268}, we can do:
\begin{Verbatim}[fontsize=\small, frame=lines]
#!/usr/bin/julia
using JuMP, MathOptAI, DecisionTree, HiGHS
truth(x::Vector) = x[1] <= 0.5 ? -2 : (x[2] <= 0.3 ? 3 : 4)
features = abs.(sin.((1:10) .* (3:4)'));
labels = truth.(Vector.(eachrow(features)));
predictor = DecisionTree.build_tree(labels, features)
model = Model(HiGHS.Optimizer);
@variable(model, 0 <= x[1:2] <= 1);
y, _ = MathOptAI.add_predictor(model, predictor, x);
\end{Verbatim}

\section{Comparison to related work}\label{sec:comparison}

Table~\ref{tab:comparison} compares the features of \MathOptAI to \OMLT \citep{omlt}, \gurobiml \citep{gurobimachinelearning}, \pyscipoptml \citep{turner2023pyscipopt}, and \gamspy \citep{gamspy}. We choose these packages because they are the most prominent and well developed in the ecosystem. Other alternatives exist---see \cite{lopezflores2024} for a comparison---but they are significantly less developed or feature-diverse than these tools. Note that the table is not intended to be a comprehensive comparison of features, for example, it does not cover alternative formulations for each predictor, supported file types for input and output, or the supported higher level interfaces to packages such as \texttt{pytorch}, \texttt{scikit-learn}, and \texttt{XGBoost}.

\begin{table}[!ht]
    \centering
    \resizebox{\textwidth}{!}{
    \begin{tabular}{r | c c c c c}
        	             & \MathOptAI & \OMLT       & \gurobiml    & \pyscipoptml & \gamspy \\
    \hline
    Programming Language & Julia      & Python      & Python       & Python       & Python \\
    License              & BSD-3      & BSD-3       & Apache-2.0   & Apache-2.0   & MIT    \\
    Modeling Language    & JuMP       & Pyomo, JuMP & gurobipy     & PySCIPOpt    & GAMS   \\
    Solvers     	     & Many       & Many        & Gurobi       & SCIP         & Many   \\
    \hline
    \textit{Formulations} \\
    Full-space           & Yes        & Yes         & Yes          & Yes          & Yes    \\
    Reduced-space        & Yes        & Yes         &              &              &        \\
    Gray-box             & Yes        &             &              &              &        \\
    GPU acceleration     & Gray-box   &             &              &              &        \\
    \hline
    \textit{Neural network layers} \\
    nn.AvgPool2D         & \revision{Yes} &         &              &              & Yes    \\
    nn.Conv2D            & \revision{Yes} & Yes     &              &              & Yes    \\
    \revisionb{nn.GCNConv} & \revisionb{Yes} & Yes  &              &              &        \\
    \revision{nn.GELU}   & \revision{Yes} &         &              &              &        \\
    \revisionb{nn.LayerNorm} & \revisionb{Yes} &      &              &              &        \\
    \revision{nn.LeakyReLU} & \revision{Yes} &      &              &              & \revision{Yes} \\
    nn.Linear            & Yes        & Yes         & Yes          & Yes          & Yes    \\
    \revision{nn.LogSoftmax} &        &             &              &              & \revision{Yes} \\
    nn.MaxPool2D         & \revision{Yes} & Yes     &              &              & Yes    \\
    nn.ReLU              & Yes        & Yes         & Yes          & Yes          & Yes    \\
   \revisionb{nn.SAGEConv} &          & Yes  &              &              &        \\
    nn.Sequential        & Yes        & Yes         & Yes          & Yes          & Yes    \\
    nn.Sigmoid           & Yes        & Yes         &              & Yes          & Yes    \\
    nn.Softmax           & Yes        & Yes         &              & Yes          & Yes    \\
    nn.Softplus          & Yes        & Yes         &              & Yes          & \revision{Yes} \\
    \revisionb{nn.TAGConv} & \revisionb{Yes} &      &              &              &        \\
    nn.Tanh              & Yes        & Yes         &              & Yes          & Yes    \\
    \hline
    \textit{Other predictor types} \\
    Binary Decision Tree & Yes        & Yes         & Yes          & Yes          & \revision{Yes} \\
    Gaussian Process     & Yes        &             &              &              &        \\
    Gradient Boosted Tree & Yes       & Yes         & Yes          & Yes          & \revision{Yes} \\
    Linear Regression    & Yes        & Yes         & Yes          & Yes          & Yes    \\
    Logistic Regression  & Yes        & Yes         & Yes          & Yes          & Yes    \\
    Random Forest        & Yes        &             & Yes          & Yes          & \revision{Yes} \\
    \end{tabular}
    }
    \caption{General comparison of software packages for embedding machine learning predictors into a mathematical optimization model. The ``Neural network layers'' section uses the PyTorch \citep{paszke2019pytorch} naming conventions.}
    \label{tab:comparison}
\end{table}

All packages are open-source. \MathOptAI is implemented in Julia, while \OMLT, \gurobiml, \pyscipoptml, and \gamspy are implemented in Python. Note that, while \gurobiml is open-source, it requires a commercial license for the Gurobi optimizer. Similarly, while \gamspy is MIT licensed, it requires a commercial license for the GAMS modeling language.
\MathOptAI, \OMLT, and \gamspy target the solver-independent modeling languages JuMP \citep{jump1}, Pyomo \citep{bynum2021pyomo}, and GAMS, while \gurobiml and \pyscipoptml are solver-specific implementations for Gurobi \citep{gurobi} and SCIP \citep{bolusani2024scip} respectively. 

\revision{\gamspy and \OMLT support} a wide variety of features and machine learning predictors. For example, \revision{\gamspy and \OMLT support} support convolution and pooling predictors, \revision{as well as a wide range of activation functions. \OMLT uniquely supports} graph neural networks (GNNs). Because \gurobiml is specific to the Gurobi optimizer, it has a smaller set of supported predictors (it is particularly missing smooth nonlinear predictors), and it implements only the full-space formulation. \gurobiml does, however, have strong support for gradient boosted trees and random forests, which coincide with Gurobi's strength as a mixed-integer optimizer. \pyscipoptml is similar to \gurobiml, except that it uses SCIP's support for global nonlinear optimization to support nonlinear neural network layers. 

\MathOptAI uniquely has support for embedding a predictor via the gray-box formulation. As a corollary, \MathOptAI is the only package to support GPU acceleration of the predictors via this formulation. In theory, the gray-box feature could be added to \OMLT, since the work of \cite{casas2025comparison} shows how to embed a PyTorch model into Pyomo. We argue that the results of \cite{parker2024formulationsscalabilityneuralnetwork} and \cite{casas2025comparison} demonstrate that the gray-box formulation is a worthwhile feature addition for \OMLT. The gray-box feature cannot be added to \gurobiml or \pyscipoptml because they require the algebraic form of the constraints for their global solvers.

Even though \OMLT can be used to build JuMP models, the user experience of \texttt{OMLT}+JuMP has more friction than \texttt{MathOptAI.jl}+JuMP because the former forces the user to install and manage both Julia and Python, and there are some nuances for when and how the \OMLT block and JuMP model interact. However, \OMLT's JuMP interface is a good option for users looking to embed predictors that are supported by \OMLT and not by \MathOptAI, such as a \revisionb{\texttt{SAGEConv} layer in a graph neural network}.

\section{Design principles}\label{sec:principles}

When designing \MathOptAI, we had the benefit of being able to observe existing implementations such as \OMLT \citep{omlt}. From this analysis, and our prior experience developing other Julia packages, we developed the following design principles.

\subsection{Leverage Python's strengths}

Julia has a strong optimization ecosystem, particularly, for our interest, around JuMP. However, Julia's machine learning ecosystem---while feature rich and well developed---is significantly less popular than Python's machine learning ecosystem. Thus, we support both Julia-based machine learning libraries such as \texttt{Lux.jl} \citep{pal2023lux} and \texttt{Flux.jl} \citep{Innes2018}, and Python-based machine learning libraries such as PyTorch \citep{paszke2019pytorch}. Our experience has shown \citep{parker2024formulationsscalabilityneuralnetwork} that the Julia-Python interface was both simple to engineer and performant to use in practice. \OMLT's similarly positive experience integrating Julia and Python libraries suggests that the future may see more integration between Python and Julia, with libraries in each language able to leverage the strengths of the other.

\subsection{Leverage Julia's strengths}

The design of \MathOptAI leverages Julia's strengths by relying heavily on multiple dispatch. Inside \MathOptAI, we define new predictors as a subtype of \texttt{AbstractPredictor}. The subtypes implement a common \texttt{MathOptAI.add\_predictor} function. For example, the basic code to implement a non-smooth nonlinear $\textsf{ReLU}$ predictor is:

\begin{Verbatim}[fontsize=\small, frame=lines]
struct ReLU <: MathOptAI.AbstractPredictor end
function MathOptAI.add_predictor(model::JuMP.Model, predictor::ReLU, x::Vector)
    y = JuMP.@variable(model, [1:length(x)])
    cons = JuMP.@constraint(model, y .== max.(0, x))
    return y, MathOptAI.Formulation(predictor, y, cons)
end
MathOptAI.add_predictor(::JuMP.Model, p::MathOptAI.ReducedSpace{ReLU}, x::Vector) =
    (max.(0, x), MathOptAI.Formulation(p))
\end{Verbatim}

A second strength of Julia is its package extension mechanism, which allows code to be dynamically loaded if two other packages are loaded. We use this feature to implement package extensions between \MathOptAI and third-party Julia packages. As one example, the code to implement the interface between the linear modeling package \texttt{GLM.jl} \citep{glm} and \MathOptAI is:

\begin{Verbatim}[fontsize=\small, frame=lines]
import MathOptAI, GLM
MathOptAI.add_predictor(model::JuMP.Model, p::GLM.LinearModel, x::Vector) =
    MathOptAI.add_predictor(model, MathOptAI.Affine(GLM.coef(p)), x)
\end{Verbatim}

The result is a clean and simple implementation, where we have a set of predictors that subtype from \texttt{AbstractPredictor} and implement a common API to convert their data structures into a set of JuMP variables and constraints, and a set of package extensions that convert third-party machine learning models into the \texttt{AbstractPredictor}s supported by \MathOptAI.

\subsection{Composition of predictors}

Packages such as \OMLT, \gurobiml, and \pyscipoptml approach neural networks with the assumption that basic building block is a layer:
$$\isvec{y} = \sigma(W\isvec{x} + \isvec{b}),$$
which combines both an affine transform and an activation function.

\MathOptAI follows the PyTorch \citep{paszke2019pytorch} convention in which everything is a predictor $\isvec{y} = F(\isvec{x}): \mathbb{R}^N \rightarrow \mathbb{R}^P$, including linear layers and nonlinear activation functions. A layer in \MathOptAI is therefore the composition of two predictors:
$$\isvec{y} = (\sigma \circ \textsf{Affine}_{(W, \isvec{b})})(\isvec{x}).$$
This approach significantly simplifies the implementation of the library because we can reuse and recombine the various predictors into different pipelines.

The complete code to implement predictor composition is the following:
\begin{Verbatim}[fontsize=\small, frame=lines]
struct Pipeline <: MathOptAI.AbstractPredictor
    layers::Vector{MathOptAI.AbstractPredictor}
end
function MathOptAI.add_predictor(model::JuMP.Model, predictor::Pipeline, x::Vector)
    formulation = MathOptAI.PipelineFormulation(predictor, Any[])
    for layer in predictor.layers
        x, inner_formulation = MathOptAI.add_predictor(model, layer, x)
        push!(formulation.layers, inner_formulation)
    end
    return x, formulation
end
\end{Verbatim}
which enables syntax like:
\begin{Verbatim}[fontsize=\small, frame=lines]
A, b = rand(2, 3), rand(2)
logistic = MathOptAI.Pipeline([MathOptAI.Affine(A, b), MathOptAI.Sigmoid()])
\end{Verbatim}
Note how \texttt{Pipeline} is itself an \texttt{AbstractPredictor}, which enables nested pipelines.

\subsection{Inputs and outputs are vectors}

A key feature of \MathOptAI are the types of inputs and outputs that are supported. One approach would be to allow the input and output types to depend on the predictor. Thus, an $\textsf{Affine}$ predictor might input and output n-dimensional arrays, while a binary classifier predictor might output a scalar Boolean. Unfortunately, such a design makes the predictor-composition principle difficult to implement.
Instead, \MathOptAI requires that all predictors accept only a vector \texttt{x::Vector} as input and return a vector \texttt{y::Vector} as output, even if the input and output is a scalar.

The biggest limitation of this approach is that \MathOptAI does not natively support $n$-dimensional array inputs or outputs. One common example is an image input, in which the natural input type is a three-dimensional array for the row, column, and color channel. \revision{Our} solution is to force users to reshape their input and output data into a one-dimensional vector. For example, a $(M, N, C)$ dimensional image can be reshaped into a vector of length $M\times N \times C$. \revision{The corresponding input size information of $(M, N, C)$ is stored as an attribute of the predictor so that we can recover the original array when needed}. This decision is a significant trade-off, but it greatly simplified the implementation of \MathOptAI and we have not found it to be a burden in practice.

\section{Conclusions}

This paper introduced \MathOptAI, an open-source Julia library for embedding trained machine learning predictors into a JuMP model. A novel feature of \MathOptAI is its support for the gray-box formulation. Readers are directed to \revision{\cite{DowsonMathOptAI} for the archived source code and to \url{https://github.com/lanl-ansi/MathOptAI.jl} for the latest version.}

The roadmap for future development of \MathOptAI includes a way to write decomposition algorithms \revision{and custom linear system solvers} that leverage a priori knowledge about the structure of the machine learning predictors. We are also interested in adding support for new package extensions and predictors; however, the space of predictors that we \textit{could} include is much larger than the set of predictors we can afford to maintain and test in an open-source package, so care is needed to avoid feature bloat.

\ACKNOWLEDGMENT{
Funding was provided by the Los Alamos National Laboratory LDRD program as part of the Artimis project.
Approved for unlimited release: LA-UR-25-24963.
}

\bibliographystyle{informs2014}
\bibliography{main}

@article{bezanson_julia_2017,
      title = {Julia: {{A Fresh Approach}} to {{Numerical Computing}}},
      volume = {59},
      number = {1},
      journal = {SIAM Review},
      author = {Bezanson, Jeff and Edelman, Alan and Karpinski, Stefan and Shah, Viral B.},
      year = {2017},
      pages = {65-98}
}

@article{lopezflores2024,
      author = {L{\'o}pez-Flores, Francisco Javier and Ram{\'i}rez-M{\'a}rquez, Cesar and Ponce-Ortega, Jose Mar{\'i}a},
      title = {Process Systems Engineering Tools for Optimization of Trained Machine Learning Models: Comparative and Perspective},
      journal = {Industrial \& Engineering Chemistry Research},
      volume = {63},
      number = {32},
      pages = {13966-13979},
      year = {2024}
}

@article{legat2021mathoptinterface,
    title={{MathOptInterface}: a data structure for mathematical optimization problems},
    author={Legat, Beno{\^\i}t and Dowson, Oscar and Dias Garcia, Joaquim and Lubin, Miles},
    journal={INFORMS Journal on Computing},
    year={2021},
    volume={34},
    number={2},
    pages={672--689}
}

@article{omlt,
    author  = {Francesco Ceccon and Jordan Jalving and Joshua Haddad and Alexander Thebelt and Calvin Tsay and Carl D Laird and Ruth Misener},
    title   = {{OMLT}: {O}ptimization \& Machine Learning Toolkit},
    journal = {Journal of Machine Learning Research},
    year    = {2022},
    volume  = {23},
    number  = {349},
    pages   = {1--8}
}

@article{jump1,
    author = {Lubin, Miles and Dowson, Oscar and Garcia, Joaquim Dias and Huchette, Joey and Legat, Beno{\^\i}t and Vielma, Juan Pablo},
    date = {2023},
    journal = {Mathematical Programming Computation},
    number = {3},
    pages = {581--589},
    title = {{JuMP} 1.0: {R}ecent improvements to a modeling language for mathematical optimization},
    volume = {15},
    year = {2023}
}

@article{huangfu_2018,
    author = {Huangfu, Q. and Hall, J. A. J.},
    title = {Parallelizing the dual revised simplex method},
    journal = {Mathematical Programming Computation}, 
    volume = {10},
    number = {1},
    year = {2018},
    pages = {119--142},
}

@article{ipopt,
  title={On the implementation of an interior-point filter line-search algorithm for large-scale nonlinear programming},
  author={W{\"a}chter, Andreas and Biegler, Lorenz T},
  journal={Mathematical Programming},
  volume={106},
  number={1},
  pages={25--57},
  year={2006},
  publisher={Springer},
}

@article{paszke2019pytorch,
  title={Pytorch: An imperative style, high-performance deep learning library},
  author={Paszke, Adam and Gross, Sam and Massa, Francisco and Lerer, Adam and Bradbury, James and Chanan, Gregory and Killeen, Trevor and Lin, Zeming and Gimelshein, Natalia and Antiga, Luca and others},
  journal={Advances in Neural Information Processing Systems},
  volume={32},
  year={2019}
}

@book{bynum2021pyomo,
    title={Pyomo--optimization modeling in {Python}}, author={Bynum, Michael L. and Hackebeil, Gabriel A. and Hart, William E. and Laird, Carl D. and Nicholson, Bethany L. and Siirola, John D. and Watson, Jean-Paul and Woodruff, David L.},
    edition={{Third}},
    volume={67},
    year={2021},
    publisher={Springer Science \& Business Media},
    address={Cham, Switzerland}
}

@article{Innes2018,
    doi = {10.21105/joss.00602},
    url = {https://doi.org/10.21105/joss.00602},
    year = {2018},
    publisher = {The Open Journal},
    volume = {3},
    number = {25},
    pages = {602},
    author = {Mike Innes},
    title = {Flux: Elegant machine learning with Julia},
    journal = {Journal of Open Source Software} 
}

@article{casas2025comparison,
title = {A comparison of strategies to embed physics-informed neural networks in nonlinear model predictive control formulations solved via direct transcription},
journal = {Computers \& Chemical Engineering},
volume = {198},
pages = {109105},
year = {2025},
author = {Carlos Andrés Elorza Casas and Luis A. Ricardez-Sandoval and Joshua L. Pulsipher},
}

@InProceedings{turner2023pyscipopt,
  title={{PySCIPOpt-ML}: Embedding trained machine learning models into mixed-integer programs},
  author={Turner, Mark and Chmiela, Antonia and Koch, Thorsten and Winkler, Michael},
  editor="Tack, Guido",
  booktitle="Integration of Constraint Programming, Artificial Intelligence, and Operations Research",
  year="2025",
  publisher="Springer Nature Switzerland",
  address="Cham",
  pages="218--234",
}

@article{bolusani2024scip,
  title={{The SCIP optimization suite 9.0}},
  author={Bolusani, Suresh and Besan{\c{c}}on, Mathieu and Bestuzheva, Ksenia and Chmiela, Antonia and Dion{\'\i}sio, Jo{\~a}o and Donkiewicz, Tim and van Doornmalen, Jasper and Eifler, Leon and Ghannam, Mohammed and Gleixner, Ambros and others},
  journal={arXiv preprint arXiv:2402.17702},
  year={2024},
  url={https://arxiv.org/abs/2402.17702}
}

@inproceedings{parker2024formulationsscalabilityneuralnetwork,
  title={{Nonlinear Optimization with GPU-Accelerated Neural Network Constraints}}, 
  author={Robert B. Parker and Oscar Dowson and Nicole LoGiudice and Manuel Garcia and Russell Bent},
  year={2025},
  booktitle={ScaleOPT: NeurIPS 2025 Workshop on GPU-Accelerated and Scalable Optimization},
  url={https://arxiv.org/abs/2509.22462}
}

@phdthesis{casas2024,
  title        = {Robust {NMPC} of Large-Scale Systems and Surrogate Embedding Strategies for {NMPC}},
  author       = {Carlos Andr{\'e}s Elorza Casas},
  year         = 2024,
  address      = {Waterloo, Ontario, Canada},
  school       = {University of Waterloo},
  type         = {Masters Thesis}
}

@misc{ben_sadeghi_2022_7359268,
  author       = {Ben Sadeghi and
                  Poom Chiarawongse and
                  Kevin Squire and
                  Daniel C. Jones and
                  Andreas Noack and
                  Cédric St-Jean and
                  Rik Huijzer and
                  Roland Schätzle and
                  Ian Butterworth and
                  Yu-Fong Peng and
                  Anthony Blaom},
  title        = {{DecisionTree.jl - A Julia implementation of the CART Decision Tree and Random Forest algorithms}},
  year         = {2022},
  publisher    = {Zenodo},
  note         = {{Version 0.11.3}},
  doi          = {10.5281/zenodo.7359268},
  url          = {https://doi.org/10.5281/zenodo.7359268}
}

@misc{pal2023lux,
  author    = {Pal, Avik},
  title     = {{Lux: Explicit Parameterization of Deep Neural Networks in Julia}},
  year      = {2023},
  publisher = {Zenodo},
  note      = {{Version 0.5.0}},
  doi       = {10.5281/zenodo.7808904},
  url       = {https://doi.org/10.5281/zenodo.7808904}
}

@misc{gurobimachinelearning,
    author={{Gurobi Optimization, LLC}},
    title={{Gurobi Machine Learning}},
    year={2026},
    url={https://github.com/Gurobi/gurobi-machinelearning},
    note={Date visited: 2026-01-01}
}

@misc{gurobi,
  author = {{Gurobi Optimization, LLC}},
  title = {{Gurobi Optimizer}},
  year = {2026},
  url = {https://www.gurobi.com},
note={Date visited: 2026-01-01}
}

@misc{gamspy,
    author={{{GAMS} Development Corporation}},
    title={{GAMSPy}: Algebraic Modeling Interface to {GAMS}},
    year={2026},
    url={https://github.com/GAMS-dev/gamspy},
    note={Date visited: 2026-01-01}
}

@misc{glm,
  author       = {Douglas Bates and
                  Andreas Noack and
                  Simon Kornblith and
                  Milan Bouchet-Valat and
                  Michael Krabbe Borregaard and
                  Alex Arslan and
                  John Myles White and
                  Dave Kleinschmidt and
                  Phillip Alday and
                  Galen Lynch and
                  Iain Dunning and
                  Patrick Kofod Mogensen and
                  Sam Lendle and
                  Dilum Aluthge and
                  Mousum Dutta and
                  pdeffebach and
                  José Bayoán Santiago Calderón, PhD and
                  Ayush Patnaik and
                  Benjamin Born and
                  Bradley Setzler and
                  Chris DuBois and
                  Jacob Quinn and
                  Ondřej Slámečka and
                  Paul Bastide and
                  Viral B. Shah and
                  Anthony Blaom, PhD and
                  Bernhard König},
  title        = {{JuliaStats/GLM.jl}},
  year         = {2023},
  publisher    = {Zenodo},
  note         = {{Version 1.9.0}},
  doi          = {10.5281/zenodo.8345558},
  url          = {https://doi.org/10.5281/zenodo.8345558}
}

@misc{DowsonMathOptAI,
  author =        {Dowson, Oscar and Parker, Robert B. and Bent, Russell},
  publisher =     {INFORMS Journal on Computing},
  title =         {{MathOptAI.jl}: Embed trained machine learning predictors into {JuMP} models},
  year =          {2025},
  doi =           {10.1287/ijoc.2025.1446.cd},
  url =           {https://github.com/INFORMSJoC/2025.1446},
  note =          {{Available} for download at \url{https://github.com/INFORMSJoC/2025.1446}},
}

@article{Albersmeyer,
author = {Albersmeyer, Jan and Diehl, Moritz},
title = {The Lifted Newton Method and Its Application in Optimization},
journal = {SIAM Journal on Optimization},
volume = {20},
number = {3},
pages = {1655-1684},
year = {2010}
}

\end{document}